\pdfoutput=1

\documentclass[11pt]{article}

\usepackage[final]{acl}
\usepackage{times}
\usepackage{latexsym}
\usepackage{hyperref}
\usepackage{float}
\usepackage{caption}
\usepackage{array} 
\usepackage{xcolor} 
\usepackage{amsmath}
\usepackage{tabularx}
\usepackage[T1]{fontenc}

\usepackage[utf8]{inputenc}

\usepackage{microtype}

\usepackage{inconsolata}

\usepackage{graphicx}

\title{UOUO: Uncontextualized Uncommon Objects for Measuring Knowledge Horizons of Vision Language Models}

\author{ 
\textbf{Xinyu Pi*}\textsuperscript{{$1$}}\quad
\textbf{Mingyuan Wu*}\textsuperscript{{$2$}}\quad
\textbf{Jize Jiang*}\textsuperscript{{$2$}}\quad
\textbf{Haozhen Zheng*}\textsuperscript{{$2$}}\\
\textbf{Beitong Tian}\textsuperscript{{$2$}}\quad
\textbf{Chengxiang Zhai}\textsuperscript{{$2$}}\quad
\textbf{Klara Nahrstedt}\textsuperscript{{$2$}}\quad
\textbf{Zhiting Hu}\textsuperscript{{$1$}} \\
{\textsuperscript{$1$}University of California San Diego} {\textsuperscript{$2$}University of Illinois Urbana-Champaign}\\
{\texttt{xpi@ucsd.edu, \{mw34, jizej2, haozhen3\}@illinois.edu}} \\
{\texttt{* indicates equal contribution}}
}
  
\begin{document}
\maketitle
\begin{abstract}
Smaller-scale Vision-Langauge Models (VLMs) often claim to perform on par with larger models in general-domain visual grounding and question-answering benchmarks while offering advantages in computational efficiency and storage. However, their ability to handle rare objects, which fall into the long tail of data distributions, is less understood. To rigorously evaluate this aspect, we introduce the "Uncontextualized Uncommon Objects" (UOUO) benchmark. This benchmark focuses on systematically testing VLMs with both large and small parameter counts on rare and specialized objects. Our comprehensive analysis reveals that while smaller VLMs maintain competitive performance on common datasets, they significantly underperform on tasks involving uncommon objects. We also propose an advanced, scalable pipeline for data collection and cleaning, ensuring the UOUO benchmark provides high-quality, challenging instances. These findings highlight the need to consider long-tail distributions when assessing the true capabilities of VLMs.
\end{abstract}

\begin{figure*}[h]
    \vspace{-8mm}
    \centering
    \includegraphics[width=\textwidth, height=6.5cm]{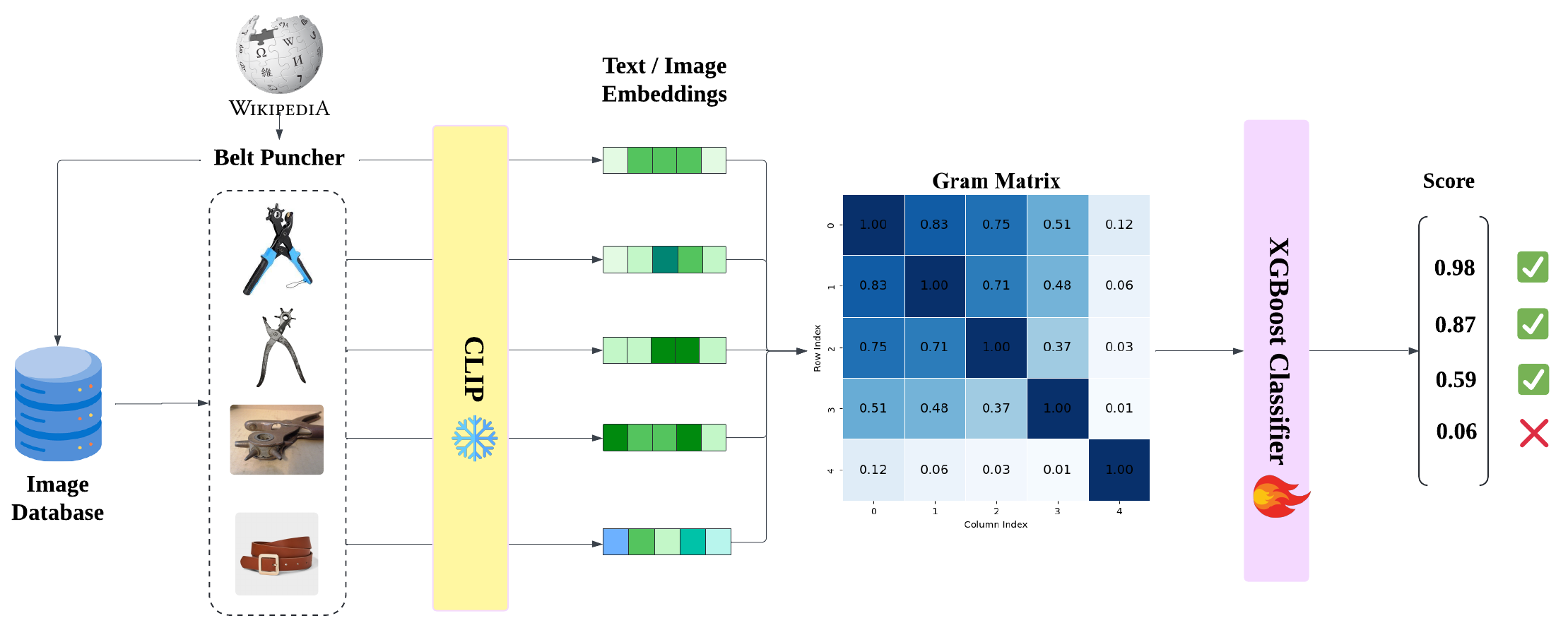} 
    \caption{UOUO Data Curation Pipeline. Snowflake means frozen weights, and fire means tune-able weights. } 
    \label{fig:uouo_workflow} 
    \vspace{-3mm}
\end{figure*}

\section{Introduction}
The advent of Vision-Language Models (VLMs) has marked a revolutionary leap in the integration of natural language processing and computer vision, largely due to the capabilities of the self-attention mechanism and the Transformer architecture \cite{vaswani2023attention}. These technologies allow VLMs to effectively process and fuse information from both text and images, leading to significant advancements in tasks that require multimodal understanding, such as visual question answering and image captioning \cite{radford2021learning, li2023blip2, alayrac2022flamingo, xu2023multimodal, young-etal-2014-image}.

VLMs, trained on large-scale datasets, typically boast high performance on general tasks involving everyday objects and common scenarios \cite{li2024llavanext-strong, du2022glm, wang2023cogvlm}. However, models of smaller scale, defined here as having fewer than 70 billion parameters, often claim to match the capabilities of their larger counterparts on general domain tasks \cite{lin2015microsoft, agrawal2016vqa, yu2016modeling, liu2024mmbench, goyal2017making, yu2023mmvet} while offering advantages in computational efficiency and storage. Despite these claims, the No-Free-Lunch Theorem \cite{wolpert1997no} suggests that these smaller models may compromise on their ability to handle less common or more complex scenarios that lie in the long tail of data distributions.

One natural and intuitive hypothesis is that they are sacrificing their fitness to the elements on the long tail of the distribution. Empirical observations of real-world data frequently align with Zipf's and Power Law \cite{piantadosi2014zipf, clauset2009power}, which indicates that while some objects and concepts are exceedingly common, a vast number of them are rare and fall into the long tail of the distribution . Understanding how well VLMs handle these rare and uncommon instances is crucial for assessing their true robustness and applicability across diverse and nuanced contexts. 

Despite the importance of this evaluation, there is currently a lack of dedicated benchmarks that systematically test VLMs on objects and concepts that are significantly outside the everyday norm. To address this gap, we introduce the "Uncontextualized Uncommon Objects" (UOUO) benchmark. The object class distribution of UOUO is systematically out of common image sources such as ImageNet \cite{russakovsky2015imagenet}, COCO \cite{lin2015microsoft}, and Open Image Dataset \cite{Kuznetsova_2020}. Our goal is to rigorously test and quantify the performance of both large-scale and small-scale VLMs on elements from the long tail of the distribution to showcase their knowledge gap.

The contribution of our work is three-fold. (1) We compile a million-scale dataset specifically designed to include uncommon and uncontextualized objects, which are rarely encountered in everyday contexts but are significant in specialized domains. (2) We evaluate the performance gap between large-scale and small-scale VLMs when dealing with these rare elements, showcasing the significant knowledge and performance gap between large- and small-scale model on the long-tail distributions. (3) We propose a systematic pipeline for automatic and scalable data collection and cleaning, ensuring high-quality and representative testing instances.

\section{Related Work}

\paragraph{Real-world VQA Benchmarks}  Based on our survey, the typical real-world visual question answering datasets (excluding mathematics, celebrity, landmark, place, OCR and chart-reading) used in popular open-source VLMs such as LLaVa \cite{li2024llavanext-strong}, CogVLM \cite{wang2023cogvlm} BLIP2 \cite{li2023blip2}, Qwen VL \cite{bai2023qwenvl} and MiniCPM-V \cite{yu2023rlhf} includes the following: COCO \cite{lin2015microsoft},
RefCOCO \cite{yu2016modeling},
NoCAPs \cite{agrawal2019nocaps},
MMBench \cite{liu2024mmbench},
VQA-v2 \cite{goyal2017making},
OK-VQA \cite{okvqa},
MME \cite{fu2024mme},
GQA \cite{hudson2019gqa}.

\textit{Much to our surprise, it turns out that the image sources of GQA, RefCoCo, OK-VQA, MME Coarse-Grained Recognition, VQA-v2, and a significant proportion of MMBench are all direct random samples from COCO.} Only NoCAPs features novel object classes (sourced from the 600-categories Open Image Dataset \cite{Kuznetsova_2020} outside COCO's less-than-100 common classes. This showcases the significant limitation of categorical diversity of extant VQA datasets. The knowledge and performance gap between the small- and large- scale VLMs might be concealed in such low coverage and diversity.

\paragraph{Existing  Datasets with Uncommon Object Labels} In extant datasets, Stanford Cars \cite{Car6755945}, CUB-bird \cite{wah2011caltech}, Deepfish \cite{saleh2020realistic}, ROCOv2 \cite{rückert2024rocov2}, FGVC-Aircraft \cite{maji13fine-grained} also features rare object labels. Some non-academic mine \& stone datasets, and chemical objects datasets can also be found on internet. However, the typical emphasis of these datasets is either \textit{fine-grained subtype} or subspecies of common objects, or \textit{domain-specific expert knowledge}. In realistic use cases such as autonomous car or embodied robotics, such knowledge might have limited generalizability.


\section{Data Curation and Filtering}

\subsection{Domain Selection and Scraping}
To construct the UOUO (Uncontextualized Uncommon Objects) benchmark, we began by selecting specific domains that are rich in specialized knowledge yet contain objects and tools that are rarely encountered by the general public. Our focus was on the industry sector, given its diversity and the presence of numerous specialized tools and equipment. These artificial tools are significantly out of the distribution of ImageNet, COCO, and Open Image Dataset. 

We used Wikipedia as a starting point, targeting the page dedicated to manufacturing (https://en.wikipedia.org/wiki/Manufacturing). For each sub-sector identified within this domain, we employed GPT-4-Turbo \cite{openai2024gpt4} to generate a list of the top 50 objects or tools pertinent to experts in the field but obscure to the general populace. This list was generated through prompt-based querying, asking the model to identify objects that are crucial within the industry but not commonly known.

Once we had our list of uncommon objects, we performed a Google Image Search for each object name. For each query, we collected the top 50 image results. This approach allowed us to gather a diverse set of images representing each object under different conditions and contexts. For detailed dataset statistics of UOUO, we refer readers to Appendix \ref{app:A}.

\paragraph{Mannual Annotation}

The image instances collected from Google Image Search can be noisy, with perhaps one fifth irrelevant instances for each queried uncommon category. To ensure the quality and relevance of the dataset, we implemented a rigorous annotation and cleaning process, combining manual and automated techniques. Our team manually reviewed the collected images for each object category to identify and remove outliers and noisy data. Categories with consistent visual representation across examples were retained, while those filled with ambiguous or irrelevant images were discarded. This initial curation aimed to maintain high fidelity to the object’s intended representation.

\paragraph{Automatic Data Cleaning}
We utilized the CLIP model to further enhance the dataset. CLIP (Contrastive Language–Image Pre-training) provides embeddings for both images and text, enabling us to compute similarities within and across categories. For each image, we extracted its CLIP image embedding \( E_i^{c} \) and the text embedding \( T_c \) of its corresponding category name \cite{radford2021learning, sun2023evaclip}.
We calculated the cosine similarity between all pairs of image embeddings within each category to construct a GRAM matrix $ G $, where $ G_{i,j} = Cosine(E_i^{c}, E_j^{c}) $. Additionally, we computed the image-text similarity for each image as $Cosine(E_i^{c}, T_c) $. Furthermore, we add basic statistical metrics, such as the percentile of the average-similarity with respect to other category members of a given instance, the mean and variance of the average-similarity of the category.

Using these computed features, we applied an XGBoost classifier to label each image instance. This classifier was trained to distinguish between high-quality and low-quality instances based on their similarity scores.

We optimized our XGBoost classifier \cite{Chen_2016} through 5-fold cross-validation and grid search to identify the best hyper-parameters. The classifier achieved an accuracy of 0.8754 on cross-validation, closely aligning with human judgment, and exhibited Macro-Average Precision, Recall, and F1-Score of 0.8631, 0.8353, and 0.8460, respectively.

\begin{figure}[t]
    \vspace{-3mm}
    \centering
    \includegraphics[width=0.5\textwidth]{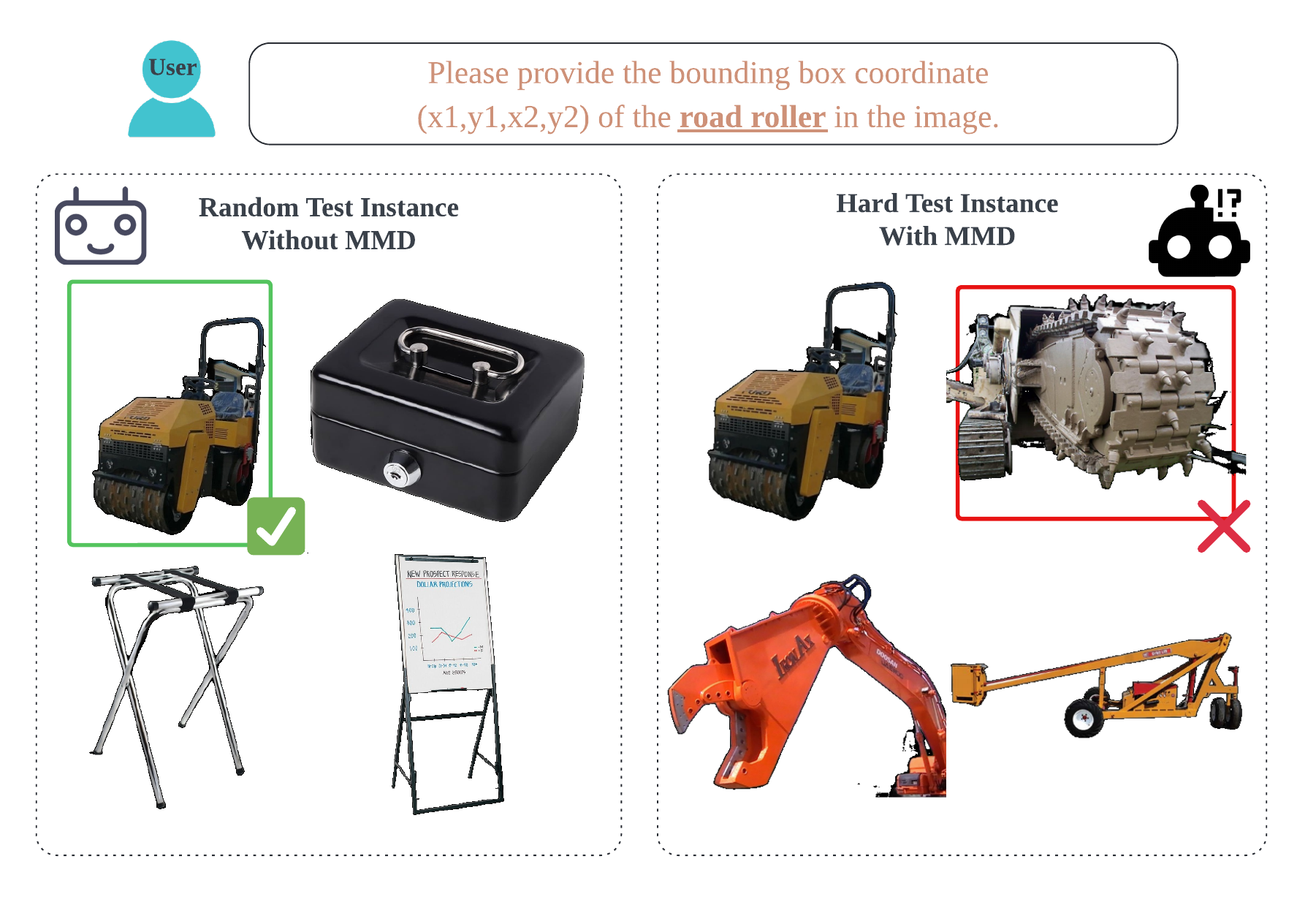} 
    \caption{With MMD, we can retrieve harder negative examples and construct higher-quality test instanecs.} 
    \vspace{-3mm}
    \label{fig:uouo_test} 
\end{figure}

\section{Test Instances Generation}

\paragraph{Background Removal and Decontextualization}
Connectionist neural networks (including VLMs) are notoriously known for their tendency of overfitting to spurious correlations present in the training data. For instance, in our  collected data, bulldozers are often seen in construction scenes laden with materials such as sand, concrete, and bricks. This high co-occurrence can lead models to rely on these contextual cues rather than truly understanding and recognizing the bulldozer itself. To mitigate this issue and ensure that models focus on the objects rather than their typical environments, we implement a robust background removal process to decontextualize all candidate objects in our dataset. To achieve effective background removal, we utilize a state-of-the-art, off-the-shelf background removal model \cite{briaai2024}.


\paragraph{Testing Instances Generation} 
To assess the performance of Vision-Language Models on our UOUO benchmark, we generated challenging test instances designed to probe the models' capabilities beyond common knowledge. Specifically, we employ the CLIP embeddings combined with the Maximum Mean Discrepancy (MMD) with a Gaussian RBF kernel \cite{dziugaite2015training} to identify and retrieve hard negative examples. 

Let \(\mathbf{x}\) and \(\mathbf{y}\) be the sets of CLIP embeddings for two different object categories, each of shape \((n, d)\), where \(n\) is the number of embeddings and \(d\) is the embedding dimension.

The Maximum Mean Discrepancy (MMD) between sets of embeddings \(\mathbf{x}\) and \(\mathbf{y}\) is calculated as follows:
$$
\text{MMD}(\mathbf{x}, \mathbf{y}) = k(\mathbf{x}, \mathbf{x}) + k(\mathbf{y}, \mathbf{y}) - 2 \cdot k(\mathbf{x}, \mathbf{y})
$$

where the Gaussian Radial Basis Function (RBF) kernel value \(k(\mathbf{a}, \mathbf{b})\) is defined as:
$$
k(\mathbf{a}, \mathbf{b}) = \frac{1}{n^2} \sum_{i=1}^n \sum_{j=1}^n \exp \left( -\frac{1}{2\sigma^2} \left\| \mathbf{a}_i - \mathbf{b}_j \right\|^2 \right)
$$
For our calculations, we set \(\sigma = 10\).

We use the Mosaic Image Augmentation Technique \cite{ge2021yolox} to generate testing data in a scalable way. Each testing data point is created from \textbf{\textit{four}} images, each background-removed. The four images contain objects of different categories but share some similar visual properties such as structures, colors, or textures. The selection of these images is determined by the Maximum Mean Discrepancy (MMD) distance between the categories they belong to. The closer the MMD distance, the more similar in features they might appear. We create an 800x800 canvas large enough to accommodate all four images. Then, each of the four images is augmented and positioned on the canvas's top-left, top-right, bottom-left, or bottom-right. The ground-truth bounding box for the object grounding is generated from the segmentation mask of background removal and normalized to be dimension-insensitive, accounting for potential differences in the VLM’s rescaling process. Figure \ref{fig:uouo_test} showcases an exemplar test instance.


\begin{table}[t]    
    \hspace{-10.5mm}
    \scriptsize 
    \begin{tabularx}{0.5461\textwidth}{|c|c|c|c|c|} 
        \hline
        \textbf{Model} & \textbf{mIoU-mmd} & \textbf{mIoU-rand} & \textbf{acc-mmd} & \textbf{acc-rand} \\
        \hline
        llava-v1.5-7b & 0.1755 & 0.4117 & 0.4160 & 0.6954 \\
        llava-v1.5-13b & 0.2334 & 0.4711 & 0.4351 & 0.7300 \\
        llava-v1.6-vicuna-7b & 0.2779 & 0.4783 & 0.4924 & 0.7511 \\
        llava-v1.6-vicuna-13b & 0.2761 & 0.4945 & 0.5220 & 0.7773 \\
        llava-v1.6-34b & 0.3774 & 0.5504 & 0.5745 & 0.8324 \\
        cogvlm-llama3-chat-19b & 0.4905 & 0.6935 & 0.4278 & 0.6024 \\
        \hline
        gemini-1.5-pro & 0.2654 & 0.2682 & 0.6326 & 0.7986 \\
        gpt-4-turbo & 0.3396 & 0.3774 & 0.6650 & 0.8970 \\
        gpt-4o & 0.3286 & 0.3472 & 0.6779 & 0.8814\\
        \hline
    \end{tabularx}
    \caption{Mosaic Grounding Performance Metrics}
    \label{tab:mosaic_grounding}
    \vspace{-3mm}
\end{table}

\section{Experiment}
\paragraph{Procedures.} Following the aforementioned test instance generation, we test both open source VLMs that are trained to perform grounding, including:
llava-v1.5-7b,
llava-v1.5-13b \cite{liu2023improvedllava},
llava-v1.6-vicuna-7b,
llava-v1.6-vicuna-13b,
llava-v1.6-34b  \cite{li2024llavanext-strong},
cogvlm-v1.5-vicuna-7b \cite{wang2023cogvlm}, and propriety VLMs including:
gemini-1.5-pro \cite{geminiteam2024gemini},
gpt-4-turbo,
gpt-4o \cite{openai2024gpt4}.

We test VLMs' performance on both randomly generated test instances and the MMD-augmented hard instances. We employ two metrics to quantitfy the performance: \textit{mIoU} - Mean IoU (Intersection over Union), a standard metric for object segmentation; and \textit{ Accuracy }, which we prompt the VLM to output one positions from "top-left, top-right, bottom-left, bottom-right", and directly evaluate whether the answer matches the ground truth.


\paragraph{Observations and Analysis.} We present all experimental results in Table \ref{tab:mosaic_grounding}. (a) Comparing horizontally across columns, we observe significant performance drops of smaller-scale models in both \textit{mIoU} and \textit{Accuracy} with the application of MMD-based hard instance generation. Notably, the performance drops of many of them are around $30\%$. This provides solid support for our initial hypothesis that smaller-scale models have some, but insufficient fitness to the long-tail distribution objects. Furthermore, the drastic performance change showcases MMD's effectiveness in generating hard instances and non-robustness of existing grounding models. (b) Comparing vertically within columns, the central tendency is that larger scale models (except Genimi which might not be trained to perform grounding) perform much better than small-scale models in accuracy. This reveals the concealed gap of knowledge horizon of small- and large- scale models, which is usually unobservable in benchmarks consist of common objects. (c) The observation that GPT-4 series can still handle the task remarkably well (near $90\%$ and $70\%$ on random and MMD settings, respectively) showcases the task's solvability, revealing the soundness of our automatically constructed test instances.
\section{Conclusion}
In our work, we introduced the UOUO benchmark to assess VLMs on objects out of everyday distributions. Our findings show that while smaller VLMs perform well on tasks of common objects, they struggle significantly with uncommon objects, unlike larger models which handle these challenges much better. This highlights the need to consider long-tail distributions in evaluations. The systematic data curation, filtering, and hard test instance generation pipeline for UOUO construction has high extensibility, paving the road of future research of long-tail distribution objects.

\clearpage

\section*{Limitations}
One limitation of our work is the reliance on automated data collection and cleaning processes, though efficient, may introduce biases or fail to capture nuanced representations compared to fully manual curation. The UOUO benchmark also currently emphasizes static images, potentially overlooking the dynamic and context-dependent nature of object recognition in real-world scenarios. Future extensions should explore a wider range of uncommon objects across various fields and consider the inclusion of video or sequential data to better reflect real-world applications. Addressing these limitations will enhance the comprehensiveness and applicability of the UOUO benchmark.

\bibliography{custom}

\appendix
\section{Appendix}
\label{app:A}
Important statistics of UOUO are listed as follows:
\begin{itemize}
    \item \textbf{Number of categories:}
    \begin{itemize}
        \item Filtered data directory: 25,864
        \item Original data directory: 27,926
        \item Percentage of categories kept: 92.6\%
    \end{itemize}
    \vspace{0.5cm}
    \item \textbf{Total number of images:}
    \begin{itemize}
        \item Filtered dataset: 678,535
        \item Original dataset: 956,167
        \item Percentage of images kept: 71.0\%
    \end{itemize}
    \vspace{0.5cm}
    \item \textbf{Images per category stats:}
    \begin{itemize}
        \item \textbf{Filtered dataset:}
        \begin{itemize}
            \item Average: 26.235
            \item Minimum: 5
            \item Maximum: 48
        \end{itemize}
        \item \textbf{Original dataset:}
        \begin{itemize}
            \item Average: 26.235
            \item Minimum: 5
            \item Maximum: 48
        \end{itemize}
    \end{itemize}
    \vspace{0.5cm}
    \item \textbf{Average percentage of images kept in each category:} 76.0\%
\end{itemize}

\begin{figure}
    \centering
    \includegraphics[height=25cm, width=10cm]{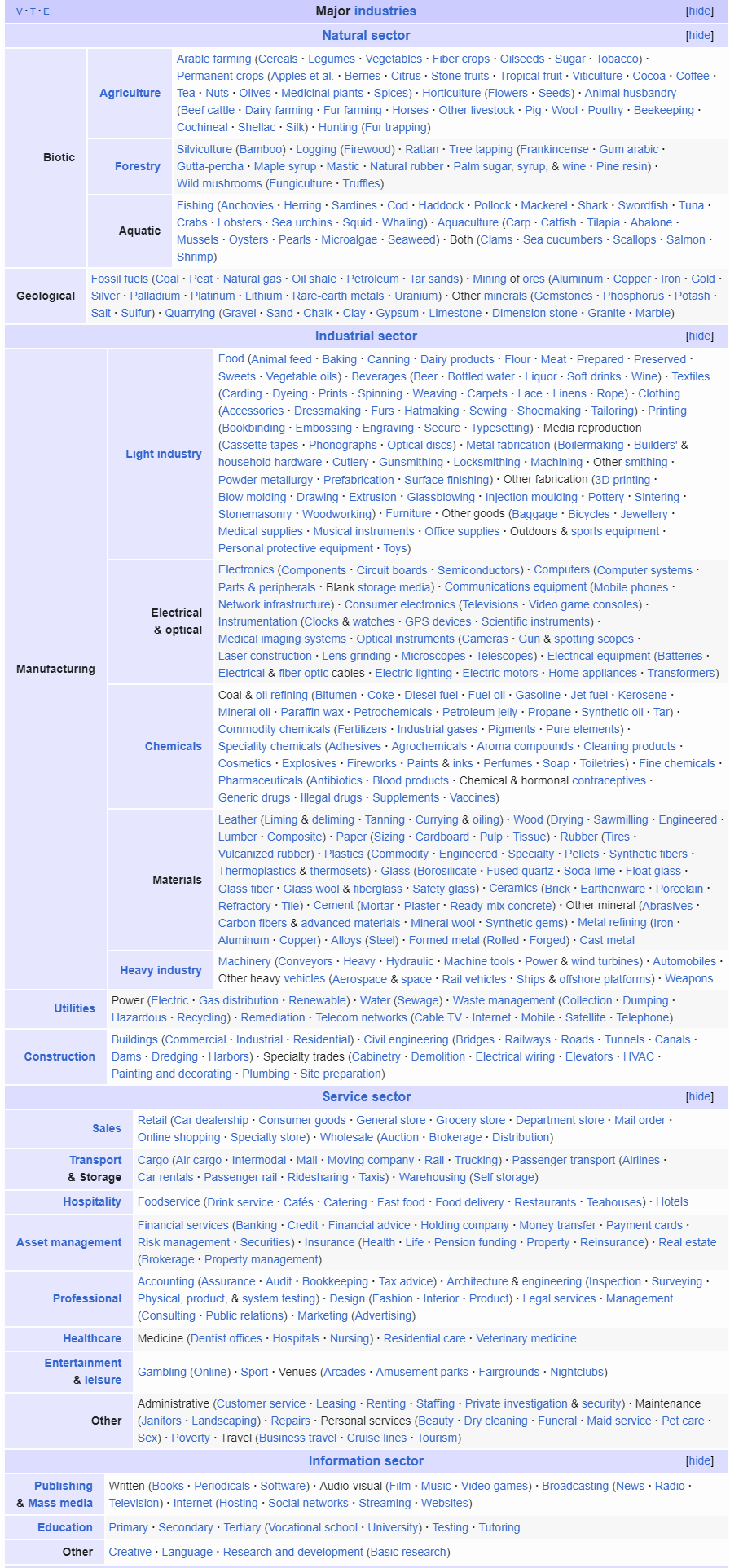}
    \caption{Wikipedia Industry List}
    \label{fig:wiki}
\end{figure}

Randomly sampled 100 categories:

\begin{table}[h!]
\centering
\begin{tabular}{|p{4cm}|p{4cm}|p{4cm}|p{4cm}|}
\hline
2D pantograph & AC Recharge Kit & Adhesive scale & Aluminum dross processing machine \\
\hline
Artificial insemination gun & Ballistic clipboard & Ballot Box (for collecting anonymous feedback) & Banjo rim lathe \\
\hline
Bingo balls & Broodstock tanks & Broom & Burnishing Stone \\
\hline
Cable Retention Sleeve & Carding Machine & Cattle Curtain & Cell Model \\
\hline
Climbing rope & Coal centrifuge & Coffee roaster & Cold Storage Backpack \\
\hline
Compressor (hardware) & Cooling Incubator & Copy Stand & Culture trays \\
\hline
Dehooking tool & Deposit Slip Printer & Disc golf basket welder & Disc repair kit \\
\hline
Display Turntables & Distillation column & Electronic rate board & Evaporating Dish \\
\hline
Extrusion laminator & Fiber disc & Fishing rod holders & Flange spreader \\
\hline
Flower press & Foundation crack ruler & Fume Extraction Hood & Goniophotometer \\
\hline
Graduated cylinders & Granule Filler & Inductively Coupled Plasma (ICP) Spectrometer & Irrigation pipelayer \\
\hline
Lacquer polishing brush & Leachate Collection Pipe & Live Feed Incubator & Longlines and ropes \\
\hline
Martingale & Metal scribe & Mobile manufacturing unit (MMU) & Mushroom grow tent \\
\hline
Music on hold player & Network Firewall Hardware & Offshore aquaculture cage & Ore skip \\
\hline
Oscillating shaker & Oxygen concentrators & Packing Gauge & Pellets coating system \\
\hline
Pellicle Formation Tool & Pillory & Pin beater & Pointer stick \\
\hline
Portable battery booster & Pressure vessels & Print Quality Inspection Scope & Pulling post \\
\hline
Purging compound dispenser & Queue stanchion & Quick release hook & Roll Coating Paint Line \\
\hline
Rope pump & Rotary drum bauxite washer & Rotary impeller feeder & Sand filter \\
\hline
Scale Breaker & Schlenk flask & Security drone & Security token device \\
\hline
Shear Line & Shock Absorber & Sign language interpreter gloves & Slab Tongs \\
\hline
Slush ice machines & Soap scum remover & Spin Welder & Spoke cutting machine \\
\hline
Spot meter & Springform pan & Tabbing shears for composite test specimens & Texture sprayer \\
\hline
Tower Climbing Harness & Violin varnish brush & Vixen Plate & Wall Hooks for Art \\
\hline
Waste basket & Water jet cutter for stone & Whalebone Scraper & Wire Mesh Cable Trays \\
\hline
\end{tabular}
\caption{List of 100 Randomly Sampled Categories}
\label{tab:sampled_categories}
\end{table}

\end{document}